\documentclass{bmvc2k}
\usepackage{multirow}
\usepackage{hhline}
\usepackage{booktabs}
\usepackage{amssymb}

\title{StereoFlowGAN: Co-training for Stereo and Flow with Unsupervised Domain Adaptation}

\addauthor{Zhexiao Xiong}{x.zhexiao@wustl.edu}{1}
\addauthor{Feng Qiao}{feng.qiao.ad@gmail.com}{2}
\addauthor{Yu Zhang}{yuzh03@gmail.com}{3}
\addauthor{Nathan Jacobs}{jacobsn@wustl.edu}{1}

\addinstitution{
 Department of Computer Science \& Engineering,\\
 Washington University in St. Louis\\
 St. Louis, MO, USA
}
\addinstitution{
 Institute for Automotive Engineering,\\
 RWTH Aachen University\\
 Templergraben 55, Aachen, Germany
}
\addinstitution{
 Department of Computer Science,\\
 University of Kentucky \\
 Lexington, KY, USA
}

\runninghead{Xiong, Qiao, Zhang, Jacobs}{StereoFlowGAN}

\begin{document}

\maketitle

\begin{abstract}
We introduce a novel training strategy for stereo matching and optical flow estimation that utilizes image-to-image translation between synthetic and real image domains. Our approach enables the training of models that excel in real image scenarios while relying solely on ground-truth information from synthetic images. To facilitate task-agnostic domain adaptation and the training of task-specific components, we introduce a bidirectional feature warping module that handles both left-right and forward-backward directions. Experimental results show competitive performance over previous domain translation-based methods, which substantiate the efficacy of our proposed framework, effectively leveraging the benefits of unsupervised domain adaptation, stereo matching, and optical flow estimation.
\end{abstract}

\section{Introduction}

Stereo matching and optical flow estimation are closely related computer vision tasks. Given an image pair, the aim of optical flow estimation is to predict a 2D vector field that reflects the pixel-wise displacement of temporally adjacent frames. In rectified stereo, the task is essentially the same. We simply use the known relative camera geometry to impose epipolar constraints, thereby reducing the correspondence-search problem from 2D to 1D. The result is a disparity map between the left image and right images which can be easily translated into a pixel-wise displacement field. We propose a co-training approach that exploits this inter-task similarity to simultaneously train networks for both tasks.

Acquiring high-quality, real training data for stereo matching and optical flow estimation is challenging and expensive, often requiring calibration between the depth sensor and stereo cameras. Therefore, optical flow estimation and stereo-matching methods highly rely on synthetic data for training. However, there is often a severe domain gap between synthetic and real data, affecting the cross-domain generalization performance. To address this problem, unsupervised domain adaptation methods have been proposed to bridge the domain gap between synthetic and real data.
 
Inspired by StereoGAN~\cite{liu2020stereogan}, which first applied domain translation to stereo matching task, we propose a multi-task framework that concurrently executes optical flow estimation and stereo matching tasks with a shared domain translation module, in which we do not need the ground-truth disparity and optical flow of target domain real images. In the shared domain translation module, we use two ResNet-based generators of opposite directions to perform cross-domain image translation. Then two discriminators are constructed to minimize the discrepancy between translated and original images. Furthermore, we adopt perceptual loss to maintain the feature-level consistency and cosine similarity loss to regularize the cross-domain generation. Utilizing synthetic2real and real images, we predict corresponding disparity maps via a stereo-matching network using only ground-truth disparity of synthetic stereo data. Simultaneously we train an optical flow estimation network using the adjacent frames of synthetic2real images and real images, leveraging ground-truth optical flow data and occlusion masks from only synthetic data. The two task-agnostic networks are jointly optimized. To connect the three modules, we build a multi-scale left-right feature warping module and a forward-backward feature warping module, which not only provide supervision for image translation but also for the training of task-specific networks. 

The key contributions of our paper are summarized below:
\begin{itemize}
\item {} We build an end-to-end joint learning framework to combine unsupervised domain translation with optical flow estimation and stereo matching in the absence of real ground truth optical flow and disparity, which facilitates the co-optimization of models, yielding superior performance compared with executing each task in isolation.
\item{} We apply novel constraints on the cycle domain translation process to achieve cross-domain translation with global and local consistency, which significantly reduces the pixel distortion during the domain translation stage.
\item{} We employ task-specific multi-scale feature warping loss and iterative feature warping loss during the training phase to regulate the training process of the shared domain translation module and task-specific module in both spatial and temporal dimensions.
\item{} Experimental results demonstrate that our proposed model achieves top-tier results compared to other unsupervised domain adaptation methods for stereo matching and optical flow estimation.
\end{itemize}

\section{Related Work}

\subsection{Stereo Matching}
The aim of stereo matching is to generate disparity maps from left and right epipolar images. Traditionally, this involved a four-step process: matching cost computation, cost aggregation, disparity computation, and disparity refinement. Since DispNet~\cite{mayer2016large}, deep learning-based works~\cite{liang2018learning, chang2018pyramid, tankovich2021hitnet, shen2021cfnet, shamsafar2022mobilestereonet} have become popular for more accurate, real-time stereo matching.

Inspired by RAFT~\cite{teed2020raft}, iterative 2D methods have been applied to this task. Notable models include AANet~\cite{xu2020aanet}, which forgoes 3D convolutions for efficiency; RAFT-Stereo~\cite{lipson2021raft}, an adaptation of previous optical flow work; CREStereo~\cite{li2022practical}, a cascaded recurrent network for practical stereo matching; and IGEV-Stereo~\cite{xu2023iterative}, which uses a combined geometry encoding volume for iterative disparity map updates.


\vspace{-0.3cm}
\subsection{Optical Flow Estimation}
Optical flow estimation aims to estimate per-pixel motion between video frames. Since FlowNet~\cite{7410673}, a series of deep neural networks have been proposed, with some~\cite{ilg2017flownet, vaquero2017joint, cheng2017segflow} using U-Net encoder-decoder structures that often lose detail in feature maps, while others~\cite{sun2018pwc, hui2018liteflownet, hui2020lightweight} use spatial pyramid networks with feature warping to reduce feature-space distance and adaptively regulate flow. 
A classic method RAFT~\cite{teed2020raft} extracts per-pixel features and iteratively updates a flow field through a recurrent unit using multi-scale 4D correlation volumes, which enables strong cross-dataset generalization. Recently, GMFlow~\cite{xu2022gmflow} and its follow-up work Unimatch~\cite{10193833}, both based on Transformers, reformulate optical flow as a global matching problem and compare feature similarities directly instead of applying extensive iterative refinements.


Besides supervised methods, some unsupervised methods~\cite{lai2019bridging, 9140315, liu2020flow2stereo, luo2021upflow, liu2020learning} have also been proposed, among which UPFlow~\cite{luo2021upflow} proposed a self-guided upsample module to tackle the interpolation blur problem in optical flow estimation, ~\cite{liu2020learning} proposed to use more reliable supervision from transformations, and ~\cite{lai2019bridging,9140315,liu2020flow2stereo} tried to utilize the relationships between stereo matching and optical flow estimation task. In our work, we suggest an efficient end-to-end co-training framework for improving performance on both tasks.

\vspace{-0.3cm}
\subsection{Unsupervised Domain Adaptation}

Transfer learning has been widely used in many computer vision tasks, such as detection~\cite{liu2022yolov5, Bu_2021_CVPR, ye2020spotpatch}, segmentation~\cite{majurski2019cell, sun2019not, xiao2022transfer}, and stereo matching~\cite{liu2022graftnet, zbontar2016stereo}. Unsupervised domain adaptation is a special transfer learning technique, which uses labeled source data and unlabeled target data, with numerous methods~\cite{tzeng2017adversarial, You_2019_CVPR, Kang_2019_CVPR, Wang_2022_CVPR1, Wang_2022_CVPR2, Ding_2022_CVPR, Huang_2022_CVPR} developed to bridge the domain gaps. Many works have applied unsupervised domain adaptation to stereo matching and optical flow estimation tasks. Key contributions include a self-adaptation method with graph Laplacian regularization~\cite{pang2018zoom}, real-time online deep stereo adaptation~\cite{tonioni2019real}, Information-Theoretic Shortcut Avoidance (ITSA)~\cite{chuah2022itsa} for domain generalization, StereoGAN~\cite{liu2020stereogan} employing an end-to-end training framework, and AdaStereo~\cite{song2021adastereo} utilizing a non-adversarial progressive color transfer algorithm. In optical flow, strategies like co-teaching~\cite{wang2022undaf} for domain alignment and meta-training~\cite{min2023meta, tonioni2019learning} have also been proposed.

\vspace{-0.2cm}
\section{Method}
\begin{figure}[t]
  \centering
  \includegraphics[width=0.7\linewidth]{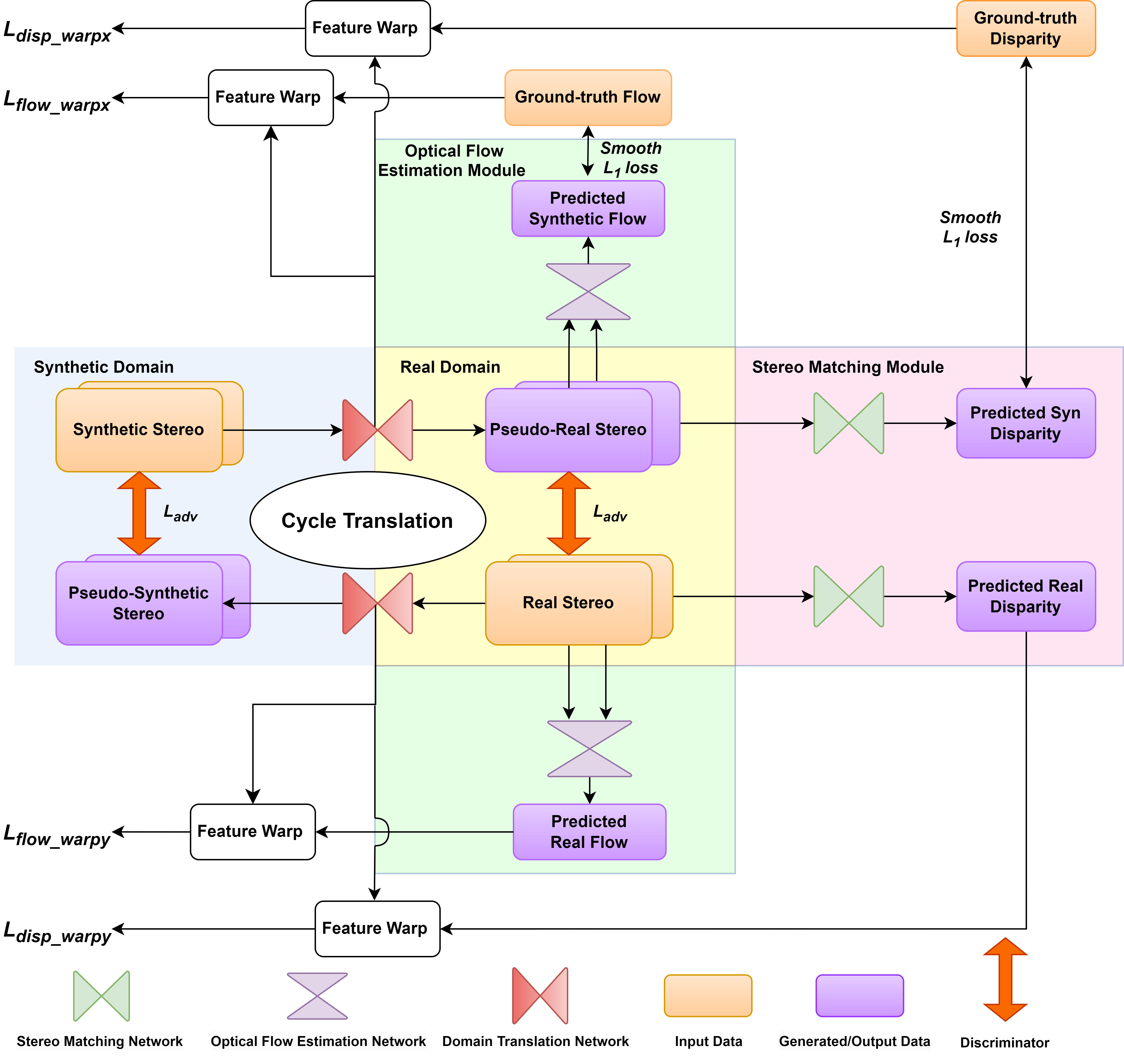}
  \caption{The framework of our proposed method. The blue block represents the synthetic domain and the yellow block represents the real domain. Domain translation is conducted between these two blocks. The pink and green blocks show stereo matching and optical flow estimation modules respectively. Please see Figure~\ref{cycle_translation} for the detail of the cycle translation module and Figure~\ref{warp} for the feature warp module.}
  \label{framework}
\end{figure}

We first describe the problem of domain translation-based optical flow estimation and stereo-matching joint training. Then we introduce the overall framework of our proposed pipeline. After that, we introduce the main components of the pipeline in detail, including the domain translation module, the stereo matching and optical flow feature warping module, and the unsupervised joint optimization scheme. The overall pipeline of our proposed framework is shown in Figure~\ref{framework}.
\subsection{Problem and Motivation}
Given a set of $N$ synthetic left-right-forward-disparity-flow tuples $\{(x_l, x_r, x
_{l\_(t+1)}, x_d, x_f)_i\}^N_{i=1}$ of source domain A, and a set of $M$ real image tuples $\{(y_l, y_r, y_{l\_(t+1)})_i\}^M_{i=1}$ of target domain B without ground truth, our goal is to conduct accurate domain translation, jointly optimize the disparity estimation network $F_{disp}$ and optical flow estimation network $F_{flow}$ for estimating the disparity $\hat{y}_d$ and optical flow $\hat{y}_f$ on the target domain. We propose to use left-right and forward-backward feature warping to jointly supervise the cross-domain translation and the task-specific framework in both spatial and temporal dimensions. 

\subsection{Domain Translation Module}
In the domain translation module, take A as the source domain and B as the target domain. In a data batch of dataloader, we load $I_{leftA}$,$I_{rightA}$, $I_{leftA\_(t+1)}$, $dispA$, $flowA$, $I_{leftB}$, $I_{rightB}$,\\$I_{leftB\_(t+1)}$. Inspired by pixel2pixel\cite{isola2017image}, we build a generator $G_{A2B}$ to translate the synthetic image $I_{leftA}$ into to real domain and get $I_{fake\_leftB}$, and a discriminator $D_B$ to help distinguish between the synthetic-to-real translated data $I_{fake\_leftB}$ and the real data $I_{leftB}$. Similarly, we build another generator $G_{B2A}$ and discriminator $D_A$ with the same structure to do real-to-synthetic image translation from $I_{leftB}$ to $I_{fake\_leftA}$ in an adversarial manner. The adversarial loss is defined as:
\begin{small}
\begin{equation}
\begin{gathered}
\mathcal{L}_{a d v}\left(G_{A2B}, D_B, \mathcal{X}, \mathcal{Y}\right)=\mathbb{E}_{y \sim\left\{\mathcal{Y}_L, \mathcal{Y}_R\right\}}\left[\log D_B(y)\right] +\mathbb{E}_{x \sim\left\{\mathcal{X}_L, \mathcal{X}_R\right\}}\left[\log \left(1-D_B\left(G_{A2B}(x)\right)\right)\right] ,
\label{L_adv}
\end{gathered}
\end{equation}
\end{small}

\noindent where $x \sim\left\{\mathcal{X}_L, \mathcal{X}_R\right\}$ represents the synthetic image pair and $y \sim\left\{\mathcal{Y}_L, \mathcal{Y}_R\right\}$ represents the real image pair. Similarly the inverse real-to-synthetic domain generation is represented as $\mathcal{L}_{a d v}\left(G_{B2A}, D_A, \mathcal{Y}, \mathcal{X}\right)$.

\begin{figure}[t]
\begin{tabular}{cc}
\begin{minipage}[t]{0.5\linewidth}
    \includegraphics[width = 1\linewidth]{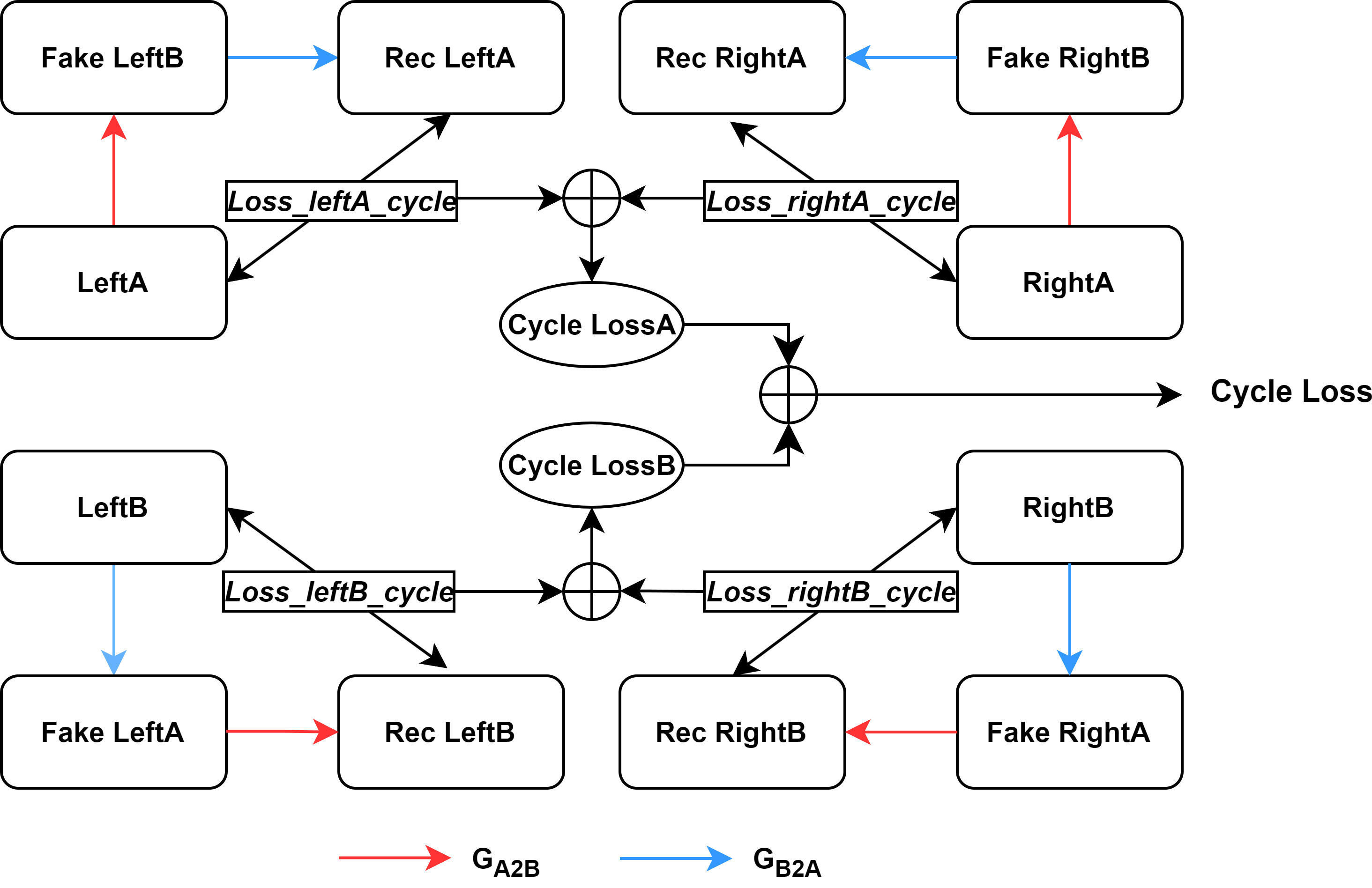}
    \caption{Cycle translation module.}
    \label{cycle_translation}
\end{minipage}
\begin{minipage}[t]{0.5\linewidth}
    \includegraphics[width = 1\linewidth]{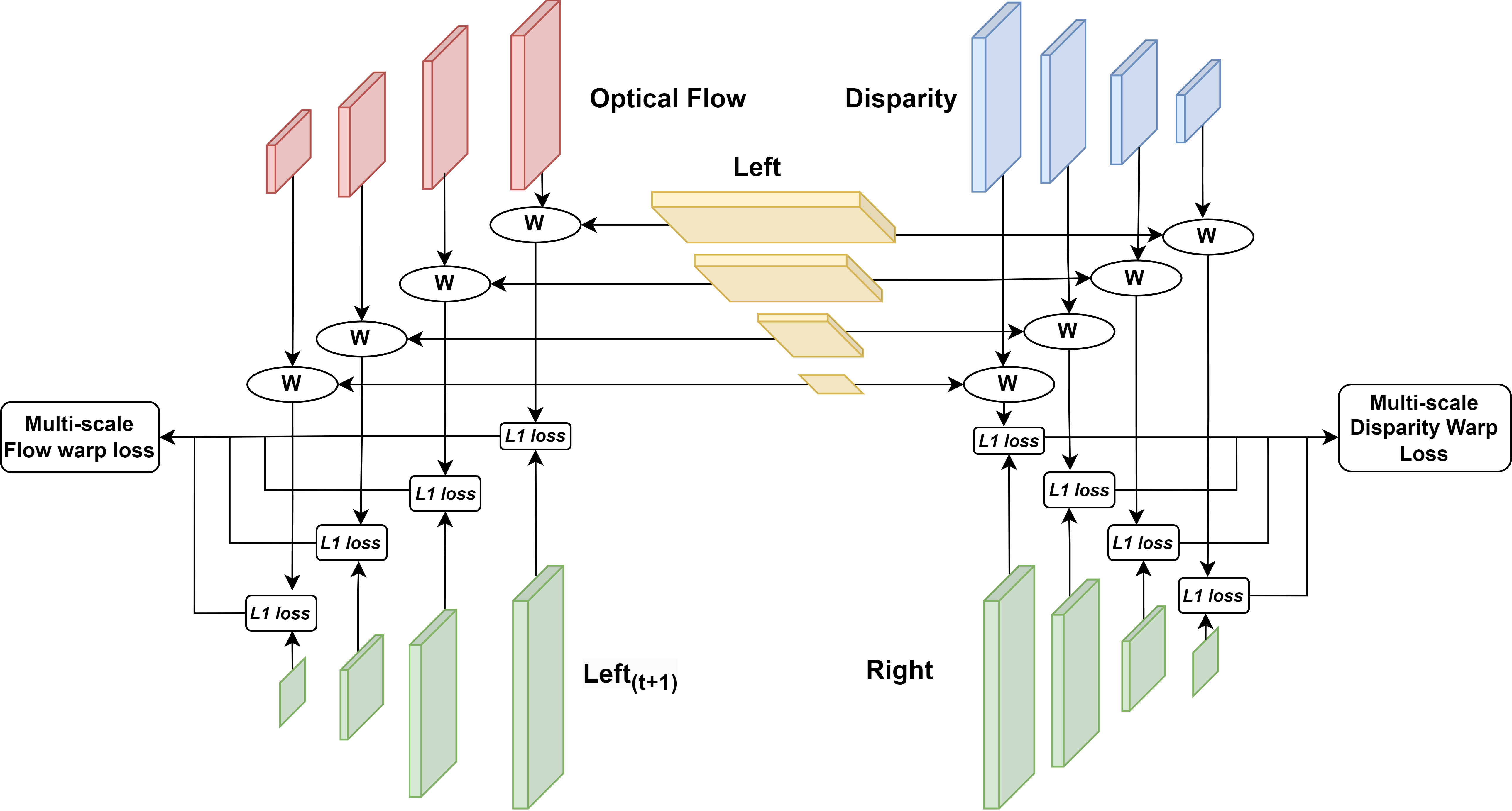}
    \caption{Multi-scale feature warping module.}
    \label{warp}
\end{minipage}
\end{tabular}
\end{figure}

Furthermore, inspired by CycleGAN~\cite{CycleGAN2017}, we make cycle domain translation from the fake-real domain back to the synthetic domain through $G_{B2A}$ and get $I_{rec\_leftA}$, with the reversed process from the fake-synthetic domain back to the real domain processed by passing $G_{A2B}$ and we get $I_{rec\_leftB}$. The framework of cycle domain translation is shown in Figure~\ref{cycle_translation}. We name it cycle loss, which is formulated as below:
\begin{small}
\begin{equation}
\begin{aligned}
& \mathcal{L}_{c y c}\left(G_{A2B}, G_{B2A}\right) = \mathbb{E}_{y \sim\left\{\mathcal{Y}_L, \mathcal{Y}_R\right\}}[\left\|G_{A2B}\left(G_{B2A}(y)\right)-y\right\|_1+(1-SSIM(G_{A2B}(G_{B2A}(y)-y)))] \\
+ & \mathbb{E}_{x \sim\left\{\mathcal{X}_L, \mathcal{X}_R\right\}}[\left\|G_{B2A}\left(G_{A2B}(x)\right)-x\right\|_1+(1-SSIM(G_{A2B}(G_{B2A}(y)-y)))],
\end{aligned}
\end{equation}
\end{small}

\noindent where $\mathbb{E}_{x \sim\left\{\mathcal{X}_L, \mathcal{X}_R\right\}}$ represents source domain image pairs and $\mathbb{E}_{y \sim\left\{\mathcal{Y}_L, \mathcal{Y}_R\right\}}$ represents target domain image pairs. $SSIM$  means structural similarity index measure(SSIM).

Specifically, to generate photorealistic cross-domain images, we use a VGG-19-based encoder-decoder structure to maintain the global feature-level similarity between the cycle-synthesized image and the source image, and adopt a perceptual loss. We also use cosine similarity to measure the pixel-level distance between the source domain and target domain and develop a cosine-similarity loss to help the domain translation network maintain local similarity, which are defined as:
\begin{small}
\begin{equation}
\begin{aligned}
& \mathcal{L}_{p}\left(G_{A2B}, G_{B2A}\right) = \mathcal{L}_{perceptual}\left(G_{A2B}\left(G_{B2A}(y)\right), y\right) +\mathcal{L}_{perceptual}\left(G_{B2A}\left(G_{A2B}(x)\right), x\right) 
\end{aligned}
\end{equation}
\end{small}
\begin{small}
\begin{equation}
\begin{aligned}
& \mathcal{L}_{cos}\left(G_{A2B}, G_{B2A}\right)=\left[1-\cos \left(G_{A2B}\left(G_{B2A}(y)\right), y\right)\right] +\left[1-\cos \left(G_{B2A}\left(G_{A2B}(x)\right), x\right)\right] 
\end{aligned}
\end{equation}
\end{small}

The domain translation loss could be calculated by summarizing the losses together in the cycle-consistency component, which is defined as:
\begin{small}
\begin{equation}
\begin{aligned}
& \mathcal{L}_{translation}\left(G_{A2B}, G_{B2A}, D_A, D_B\right)=\mathcal{L}_{a d v}\left(G_{A2B}, D_B, \mathcal{X}, \mathcal{Y}\right) +\mathcal{L}_{a d v}\left(G_{B2A}, D_A, \mathcal{Y}, \mathcal{X}\right) \\
&+\lambda_{c y c} \mathcal{L}_{c y c}\left(G_{A2B}, G_{B2A}\right)+\mathcal{L}_{p}\left(G_{A2B}, G_{B2A}\right) +\mathcal{L}_{cos}\left(G_{A2B}, G_{B2A}\right)
\label{L_translation}
\end{aligned}
\end{equation}
\end{small}

\vspace{-0.4cm}
\subsection{Stereo Matching and Optical Flow Feature Warping Loss}

Feature warping losses are widely used in stereo matching and optical flow estimation tasks. Direct domain translation may lack the precise location information between the left and right images. Therefore, to supervise the training of domain translation and help generate images with precise information, we extract features of the source image and conduct feature warping between both left-right images and forward-backward images. For models like DispNetC, as they naturally output multi-scale disparities from correlation features of network layers, we do multi-scale warp in the real-synthetic-real cycle translation. For recent models like IGEV-Stereo~\cite{xu2023iterative}, and Unimatch~\cite{10193833}, as they are trained in an iterative refinement method like RAFT~\cite{teed2020raft}, we extract the predicted disparity map or optical flow of different refinement stages. Inspired by \cite{lipson2021raft}, we calculated the smooth $\mathcal{L}_1$ loss between the warped image and target image during different refinement stages, and calculate the weighted warping loss. The framework of warping loss is shown in Figure~\ref{warp}.

\paragraph{Synthetic Image Loss}
During the training of the domain translation network, if the generators are well-trained, based on the synthetic ground-truth disparity and optical flow, the warped features should match the features of the target image exactly. Therefore, to supervise the training of the generators, we warp the feature maps of $G_{A2B}$ and $G_{B2A}$ in the synthetic-real-synthetic cycle translation process. The left-right warping loss for synthetic images is formulated as Eq~\ref{L_trans1}.
\vspace{-0.3cm}
\begin{small}
\begin{equation}
\begin{aligned}
& \mathcal{L}_{disp\_warpx}\left(G_{A2B}, G_{B2A}\right) = \mathbb{E}_{\left(x_l, x_r, x_d\right) \sim \mathcal{X}} \frac{1}{T_1} \sum_{i=1}^{T_1}\left[\left\|W\left(G_{A2B}^{(i)}\left(x_l\right), x_d\right)-G_{A2B}^{(i)}\left(x_r\right)\right\|_1\right. \\
+ & \left.\left\|W\left(G_{B2A}^{(i)}\left(G_{A2B}\left(x_l\right)\right), x_d\right)-G_{B2A}^{(i)}\left(G_{A2B}\left(x_r\right)\right)\right\|_1\right],
\label{L_trans1}
\end{aligned}
\end{equation}
\end{small}

\noindent in which $T_1$ is the number of extracted feature layers of the domain translation generator for the stereo matching task. $G^{(i)}(x)$ represents the feature of image $x$ at $i$th-layer in the domain translation network G, the warping function $W(G^{(i)}(x_l),x_d)$ warps the left feature map $G^{(i)}(x_l)$ with the ground truth disparity $x_d$. The inverse warp from $I_{rightA}$ to $I_{fake\_leftB}$ is conducted in the meantime.
Similarly, we use the forward-backward warping loss to provide temporal supervision. Based on the predicted flow, We use $G_{A2B}$ generator to warp the image from $t$ time synthetic domain $I_{leftA\_t}$ to $(t+1)$ time real domain $I_{fake\_leftB\_(t+1)}$, and conduct an inverse warp from $(t+1)$ time synthetic domain $I_{leftA\_(t+1)}$ to $t$ time real domain $I_{fake\_leftB\_t}$ in the meantime. The process is formulated as:
\begin{small}
\begin{equation}
\begin{aligned}
& \mathcal{L}_{flow\_warpx}\left(G_{A2B}, G_{B2A}\right) = \mathbb{E}_{\left(x_l, x_{l\_(t+1)}, x_f\right) \sim \mathcal{X}} \frac{1}{T_2} \sum_{i=1}^{T_2}\left[\left\|W\left(G_{A2B}^{(i)}\left(x_l\right), x_f\right)-G_{A2B}^{(i)}\left(x_{l\_(t+1)}\right)\right\|_1\right. \\
+ & \left.\left\|W\left(G_{B2A}^{(i)}\left(G_{A2B}\left(x_l\right)\right), x_f\right)-G_{B2A}^{(i)}\left(G_{A2B}\left(x_{l\_(t+1)}\right)\right)\right\|_1\right],
\label{L_trans2}
\end{aligned}
\end{equation}
\end{small}

\noindent in which $T_2$ is the number of extracted feature layers of the domain translation generator for the optical flow estimation task and $x_f$ is the ground truth flow of $t$ time left image. The feature warping loss of synthetic images serves as a bond between the shared domain translation module and the task-specific module, which supervises the generator to maintain feature-level consistency in the domain translation process. 
\vspace{-0.2cm}
\paragraph{Real Image Loss}
During the training of the task-specific modules, we further use multi-scale disparity maps and flow to warp the feature maps of $G_{B2A}$. If the task-specific networks are well-trained, based on the estimated disparity and flow of real data, the warped feature maps should match the feature maps of the target images. Therefore, we conduct left-right and forward-backward feature warping to supervise the training of stereo matching and optical flow estimation modules respectively. For the stereo matching task, based on the predicted disparity, we extract multi-scale features of the image and use the $G_{B2A}$ generator to warp the image from $I_{fake\_rightA}$ to $I_{leftB}$. In the meantime, we also conduct inverse warp from $I_{fake\_leftA}$ to $I_{rightB}$. The disparity warping loss of real images is defined as:
\begin{small}
\begin{equation}
\begin{aligned}
& \mathcal{L}_{disp\_warpy}(G_{B2A}) = \mathbb{E}_{(y_l, y_r) \sim (\mathcal{Y}_L, \mathcal{Y}_R)} \frac{1}{T_1} \sum_{i=1}^{T_1}\left[\left\|W\left(G_{B2A}^{(i)}(y_{l}), \hat{y}_d\right)-G_{B2A}^{(i)}(y_r)\right\|_1\right. \\
& + \left.\left\|W\left(G_{B2A}^{(i)}(y_r), -\hat{y}_d\right)-G_{B2A}^{(i)}(y_l)\right\|_1 \right] ,
\end{aligned}
\end{equation}
\end{small}
\noindent where $\hat{y}_d$ is the estimated disparity of real stereo image pairs by $F_{disp}(y_l,y_r)$.

Similarly, for the optical flow estimation task, we use the forward-backward warping loss to provide supervision and help maintain temporal consistency. Based on the predicted flow, We use $G_{B2A}$ generator to warp the image from $t$ time real domain $I_{leftB\_t}$ to $(t+1)$ time synthetic domain $I_{fake\_leftA\_(t+1)}$, and similarly do an inverse warp from $(t+1)$ time real domain $I_{leftB\_(t+1)}$ to $t$ time synthetic domain $I_{fake\_leftA\_t}$. The flow warping loss of real images is formulated as:
%
\begin{small}
\begin{equation}
\begin{aligned}
& \mathcal{L}_{flow\_warpy}(G_{B2A}) = \mathbb{E}_{(y_t, y_{t+1}) \sim (\mathcal{Y}_t, \mathcal{Y}_{t+1})} \frac{1}{T_2} \sum_{i=1}^{T_2}\left[\left\|W\left(G_{B2A}^{(i)}(y_{t}), \hat{y}_f\right)-G_{B2A}^{(i)}(y_{t+1})\right\|_1\right. \\
& + \left.\left\|W\left(G_{B2A}^{(i)}(y_{t+1}), -\hat{y}_f\right)-G_{B2A}^{(i)}(y_{t})\right\|_1 \right] ,
\end{aligned}
\end{equation}
\end{small}

\noindent where $\hat{y}_f$ is the estimated optical flow of $t$ time real stereo image pairs by $F_{flow}(y_t,y_{t+1})$.

\subsection{Stereo Matching and Optical Flow joint training}
Based on the cross-domain synthesized images, we jointly train stereo matching and optical flow estimation networks. We calculate the smooth $L_1$ loss between the predicted disparity/flow and estimated disparity/flow, during which features in the different refinement stages are all used under the supervision of the refinement stage. The loss functions are summarized  below:
\begin{small}
\begin{equation}
\mathcal{L}_{disp}(F_{disp})=\mathbb{E}_{\left(x_l, x_r, x_d\right) \sim \mathcal{X}}\left[\left\|F_{disp}\left(G_{A2B}\left(x_l\right), G_{A2B}\left(x_r\right)\right)-x_d\right\|_1\right] ,
\end{equation}
\begin{equation}
\mathcal{L}_{flow}(F_{flow})=\mathbb{E}_{\left(x_t, x_{t+1}, x_f\right) \sim \mathcal{X}}\left[\left\|F_{flow}\left(G_{A2B}\left(x_t\right), G_{A2B}\left(x_{t+1}\right)\right)-x_f\right\|_1\right] ,
\end{equation}
\end{small}

\noindent where $F_{disp}$ is the stereo matching network for estimating disparity and $F_{flow}$ is the optical flow estimation network for estimating optical flow from real domain stereo images of left-right views and forward-backward views. We try different stereo-matching and optical flow estimation networks to evaluate the effectiveness of our proposed framework.  
\vspace{-0.2cm}
\subsection{Joint Optimization}
In the training process, we train the domain translation module, the stereo matching module, and the optical flow estimation module in an iterative way. For every $k$ iteration, we update the gradient of the domain translation module while freezing the weights of stereo matching and optical flow estimation networks. During the $nk$ to $(n+1)k-1$ iterations, the gradients of stereo matching and optical flow estimation modules are updated in the meantime while the parameters of the domain translation module are frozen.

Besides the losses we introduced above, we borrow correlation consistency loss $\mathcal{L}_{corr}$ and mode-seeking loss $\mathcal{L}_{ms}$ from StereoGAN~\cite{liu2020stereogan}, and follow the same loss setting as this work. 
The total loss for the domain translation network is the weighted sum of the individual loss functions:
%
\begin{small}
\begin{equation}
\begin{aligned}
& \mathcal{L}_T(G_{A2B}, G_{B2A}, D_A, D_B) = \mathcal{L}_{translation}(G_{A2B}, G_{B2A}, D_A, D_B) + \lambda_{f_{disp\_warpx}} \mathcal{L}_{f_{disp\_warpx}}(G_{A2B}, G_{B2A}) \\
& + \lambda_{f_{flow\_warpx}} \mathcal{L}_{f_{flow\_warpx}}(G_{A2B}, G_{B2A}) + \lambda_{corr} \mathcal{L}_{corr}(G_{A2B}, G_{B2A}) + \lambda_{ms} \mathcal{L}_{ms}(G_{A2B})
\end{aligned}
\end{equation}
\end{small}

\noindent For the stereo matching network, the loss is formulated as:
\begin{small}
\begin{equation}
\begin{aligned}
& \mathcal{L}_d(F_{disp}, G_{B2A}) = \lambda_{disp} \mathcal{L}_{disp}(F_{disp}) + \lambda_{f_{disp\_warpy}} \mathcal{L}_{f_{disp\_warpy}}(G_{B2A})
\end{aligned}
\end{equation}
\end{small}

\noindent For the optical-flow estimation task, the loss is formulated as:
\begin{small}
\begin{equation}
\begin{aligned}
& \mathcal{L}_f\left(F_{flow},  G_{B2A}\right) =  \lambda_{flow} \mathcal{L}_{flow}(F_{flow}) + \lambda_{f_{flow\_warpy}} \mathcal{L}_{f_{flow\_warpy}}(G_{B2A}) 
\end{aligned}
\end{equation}
\end{small}

In the equations above, $\lambda_s, s \in\{translation, f_{disp\_warpx}, f_{flow\_warpx}, corr,ms,disp,  f_{disp\_warpy},\\ flow, f_{flow\_warpy} \}$ represents the weights of the different losses respectively. In the training stage, we jointly optimize $\mathcal{L}_T$, $\mathcal{L}_d$ and $\mathcal{L}_f$ together.

\begin{table}[t]
\centering
\caption{Results on datasets from Driving to KITTI2015. We take IGEV-Stereo~\cite{xu2023iterative} as the stereo matching network and Unimatch-flow~\cite{10193833} as the optical-flow estimation network. Source only means training on Driving and directly fine-tuning on KITTI2015.}
\footnotesize
\begin{tabular}{@{}cccccccc@{}}
\toprule
Method                  & EPE    & D1-all(\%) & \textgreater{}2px(\%) & \textgreater{}4px(\%) & \textgreater{}5px(\%) & EPE(flow) & F1-all(\%) \\ \midrule
IGEV-Stereo source only & 2.48   & 16.40   & 28.23                 & 11.08                 & 8.06                  & $-$    & $-$      \\
Stereo GAN ~\cite{liu2020stereogan}   & 1.65   & 10.55  & 18.59                 & 7.57                  & 5.90                   & $-$    & $-$      \\
Unimatch-flow source only      & $-$ & $-$ & $-$                & $-$                & $-$                & 14.72     & 42.20        \\
proposed   & \textbf{1.56}   & \textbf{9.16}   & \textbf{16.29}     & \textbf{6.48}                  & \textbf{4.94}       & \textbf{7.20}      & \textbf{29.48}       \\ \bottomrule
\end{tabular}
\label{driving1}
\end{table}

\begin{table}[t]
\centering
\caption{Results on datasets from Driving to KITTI2015. We take DispNetC ~\cite{mayer2016large} as the stereo-matching network and Unimatch-flow as the optical-flow estimation network.}

\footnotesize
\begin{tabular}{@{}cccccccc@{}}
\toprule
Method                   & EPE    & D1-all(\%) & \textgreater{}2px(\%) & \textgreater{}4px(\%) & \textgreater{}5px(\%) & EPE(flow) & F1-all(\%) \\ \midrule
DispNetC source only      & 7.56   & 53.84  & 65.91                 & 45.05                 & 38.36                 & $-$    & $-$      \\
Stereo GAN~\cite{liu2020stereogan}               & 3.65   & 36.36  & 51.31                 & 27.24                 & 20.79                 & $-$    & $-$      \\
Unimatch-flow source only       & $-$ & $-$ & $-$                & $-$                & $-$                & 14.72     & 42.20        \\
proposed  & \textbf{2.98}   & \textbf{29.62}  & \textbf{44.31 }     & \textbf{21.52}                 & \textbf{16.13}                 & \textbf{8.30}       & \textbf{28.79}       \\ \bottomrule
\end{tabular}
\label{driving2}
\end{table}

\begin{table}[t]
\centering
\caption{Results on datasets from VKITTI2 to KITTI2015. We take IGEV-Stereo~\cite{xu2023iterative} as the stereo matching network and Unimatch-flow~\cite{10193833} as the optical-flow estimation network. Source only means training on VKITTI2 and directly fine-tuning on KITTI2015.}
\footnotesize
\begin{tabular}{@{}cccccccc@{}}
\toprule
Method                  & EPE    & D1-all(\%) & \textgreater{}2px(\%) & \textgreater{}4px(\%) & \textgreater{}5px(\%) & EPE(flow) & F1-all(\%) \\ \midrule
IGEV-Stereo source only & 1.01   & 3.80   & 7.91  & 2.78  & 2.23   & $-$    & $-$      \\
Stereo GAN ~\cite{liu2020stereogan}   & 0.98   & 3.59  & 7.52  & 2.67  & 2.13 & $-$    & $-$      \\
Unimatch-flow source only      & $-$ & $-$ & $-$   & $-$    & $-$  & 5.79  & 21.81        \\
proposed   & \textbf{0.93}   & \textbf{3.18}   & \textbf{7.04}     & \textbf{2.37}                  & \textbf{1.90}       & \textbf{5.19}      & \textbf{18.32}       \\ \bottomrule
\end{tabular}
\label{VKITTI-IGEV}
\end{table}

\begin{table}[ht]
\centering
\caption{Results on datasets from VKITTI2 to KITTI2015. We take DispNetC~\cite{mayer2016large} as the stereo-matching network and Unimatch-flow as the optical-flow estimation network.}

\footnotesize
\begin{tabular}{@{}cccccccc@{}}
\toprule
Method                   & EPE    & D1-all(\%) & \textgreater{}2px(\%) & \textgreater{}4px(\%) & \textgreater{}5px(\%) & EPE(flow) & F1-all(\%) \\ \midrule
DispNetC source only      & 1.30   & 7.14  & 14.27    & 4.75  & 3.45 & $-$    & $-$      \\
Stereo GAN~\cite{liu2020stereogan}   & 1.27   & 6.78  & 13.02    & 4.70  & 3.50                 & $-$    & $-$      \\
Unimatch-flow source only       & $-$ & $-$ & $-$                & $-$                & $-$                & 5.79     & 21.81        \\
proposed  & \textbf{1.18}   & \textbf{5.98}  & \textbf{11.83}     & \textbf{4.12}                 & \textbf{3.03}     & \textbf{4.98}       & \textbf{19.30}       \\ \bottomrule
\end{tabular}
\label{VKITTI-dispnet}
\end{table}


\begin{table}[ht]
\centering
\caption{Ablation study on datasets from Driving to KITTI2015 with different objectives. Lower value means better performance.}
\footnotesize
\begin{tabular}{@{}c|ccccc@{}}
\toprule
\multicolumn{1}{l|}{Model}                 & Method                   & EPE  & D1-all(\%) & EPE(flow) & F1-all(\%) \\ \midrule
\multirow{5}{*}{DispNet+ Unimatch-flow}    & baseline~\cite{liu2020stereogan} & 3.65 & 36.36      & $-$    & $-$      \\
                                           & w/ perceptual loss        & 3.46 & 33.16      & $-$    & $-$      \\
                                           & w/ cosine similarity loss & 3.48 & 32.45      & $-$    & $-$      \\
                                           & w/o flow warp            & 3.06 & 31.06      & 11.40      & 37.56       \\
                                           & full obj.             & \textbf{2.98} & \textbf{29.62}      & \textbf{8.30}     & \textbf{28.79}       \\ \midrule
                                           
\multirow{5}{*}{IGEV-Stereo+Unimatch-flow} & baseline                 & 1.65 & 10.55      & $-$    & $-$      \\
                                           & w/ perceptual loss        & 1.61 & 10.03      & $-$    & $-$      \\
                                           & w/ cosine similarity loss & 1.62 & 9.97       & $-$    & $-$      \\
                                           & w/o flow warp     & 1.60   & 9.57           & 12.22      & 37.96       \\
                                           & full obj.          & \textbf{1.56} & \textbf{9.16}       & \textbf{7.20}      & \textbf{29.48}       \\ \bottomrule

\end{tabular}
\label{ablation}
\end{table}

\vspace{-0.8cm}
\begin{table}[ht]\scriptsize%
\centering
\setlength{\abovecaptionskip}{0.2cm}
\caption{Evaluation of Real-Synthetic-Real Cycle Translation}
\setlength{\tabcolsep}{1.0mm}{
\begin{tabular}{@{}ccccccc@{}}
\toprule
      & CycleGAN & w/disp\_warpx & w/$L_{per}$ & w/$L_{cos}$ & w/flow\_warpx & \textbf{Ours(full)} \\ \midrule
PSNR$\uparrow$  & 20.42         &  23.09     &  22.96  &  23.77      &  23.97     &  \textbf{24.50}                  \\
SSIM$\uparrow$  & 0.8710         &  0.8725   &   0.9076 &  0.8802     &  0.9148    &  \textbf{0.9355}                  \\
LPIPS$\downarrow$ & 0.2850        & 0.2545     & 0.1588   & 0.2053     &  0.1404    & \textbf{0.1018}                   \\ \bottomrule
\end{tabular}}
\label{ablation1}
\end{table}

\vspace{0.3cm}
\section{Experiments}
In this section, we validate the effectiveness of our proposed method for unsupervised learning of stereo matching and optical flow estimation on several standard benchmark datasets.

\vspace{-0.3cm}
\subsection{Datasets}
We implement our experiments on three autonomous driving datasets. The first one is Driving, which is a subset of a commonly used synthetic dataset, Sceneflow~\cite{MIFDB16}. The sum of this subset is 4400 in total, with both ground-truth disparity and optical flow provided. The image size is $540 \times 960$ and the disparity value range from 0 to 300. The second one is Virtual-KITTI2~(VKITTI2)~\cite{cabon2020virtual} dataset, which is a large-scale virtual autonomous driving dataset with rich weather conditions. The resolution of the images is $1920\times1080$. The third one is the widely used KITTI2015 dataset, including 200 training images collected in real scenarios and the image size is $375\times1242$. We split the training set into 160 images for training and 40 images for validation. During the training stage, we use Driving and VKITTI2 datasets as synthetic data and the 160-image split from KITTI2015 as real data, and report the performance on the 40-image validation split.

\vspace{-0.2cm}
\subsection{Evaluation Metrics}
For stereo matching task, we use the standard End-Point Error~(EPE) and D1-all metrics to evaluate the performance of the model, among which EPE is the mean average disparity error in pixels, and D1-all means the percentage of pixels whose absolute disparity error is larger than 3 pixels or 5\% of ground-truth. Also, we report the percentages of erroneous pixels larger than 2, 4, and 5. For the optical-flow estimation task, besides using EPE, we also use percentage or erroneous pixels~(F1-all) as evaluation metrics, which share the same definition as D1-all.

\vspace{-0.3cm}
\subsection{Experimental Details}
We implement our algorithm using Pytorch with Adam optimizer and AdamW optimizer for stereo matching and optical flow estimation networks respectively. We scale the images to the resolution of $512 \times 256$ during the training stage. For a fair comparison with the previous GAN-based stereo matching method StereoGAN~\cite{liu2020stereogan}, we do not use data augmentation in our training stage. We empirically set the weight factors of the losses as $\lambda_{translation} = 10$, $\lambda_{f_{disp\_warpx}}=5$, $\lambda_{f_{flow\_warpx}}=5$, $\lambda_{corr}=1$, $\lambda_{ms}=0.1$, $\lambda_{disp}=1$, $\lambda_{f_{disp\_warpy}}=5$, $\lambda_{flow}=1$, $\lambda_{f_{flow\_warpy}}=5$.

\vspace{-0.2cm}
\subsection{Results compared with other methods}
We compare our method with other methods on both domain adaptation based unsupervised stereo matching and unsupervised optical flow estimation tasks. We use Driving \& KITTI2015 and VKITTI2 \& KITTI2015 as our datasets, which are shown in Table~\ref{driving1} \&~\ref{driving2} and Table~\ref{VKITTI-IGEV} \&~\ref{VKITTI-dispnet} respectively. For a fair comparison, for the stereo matching task, we compare our method with StereoGAN, which is the only previous work that applies image-to-image domain translation into stereo matching, and we do not use data augmentation in the training process. Notice that for the Driving dataset, we use frames\_finalpass data, which is harder than frames\_cleanpass used by StereoGAN. For IGEV-Stereo and Unimatch-flow joint training, we compare the results with the source-only result on these tasks, which are trained only on the source dataset and directly tested on the KITTI2015 validation set. 

As VKITTI2 dataset contains complex autonomous driving scenes, like different weather conditions (fog, clouds, and rain) and times of day (morning and sunset), there is a larger domain gap between the source and target datasets. Therefore, the improvement from VKITTI2 to KITTI2015 is not as significant as from Driving to KITTI2015. Please see the visualization in the supplementary material.

\subsection{Ablation Study}
We conduct experiments to evaluate the efficiency of the loss functions to improve the effect of domain translation and the multi-scale warping loss of optical flow estimation and stereo matching, which is shown in Table~\ref{ablation}. The perceptual loss and cosine similarity loss help the domain translation network generate images with both global and local consistency, which contribute to the training of stereo matching and optical flow estimation networks. In the training stage, we find that the feature warping loss serves as strong supervision, which not only contributes to better performance in evaluation but also contributes to the convergence of the optical flow estimation networks, which demonstrates the effectiveness of our proposed framework. 

We also conduct further experiments to get the quantitative results of domain translation and compare it with CycleGAN~\cite{CycleGAN2017} and StereoGAN~\cite{liu2020stereogan}. We calculate the PSNR, SSIM and LPIPS between the real-synthetic-real translated image and the ground truth real image of our validation set, which reflects the domain translation ability of our model on both translation directions, shown in Table~\ref{ablation1}. It shows that our method improves the quality of domain translation.

\vspace{-0.3cm}
\section{Conclusion}

We proposed a novel co-training framework that combines domain translation, stereo matching, and optical flow estimation. We demonstrated that models trained using our framework, which incorporates a multi-scale feature warping and a cycle-consistency loss, achieve better performance on both stereo matching and optical flow estimation tasks. The strong performance of our models on real images, all without any ground-truth labels for real images, demonstrates the effectiveness of our proposed framework in bridging the domain gap between the synthetic and real data domains.


\bibliography{egbib}

\end{document}